# A Superimposed Divide-and-Conquer Image Recognition Method for SEM Images of Nanoparticles on The Surface of Monocrystalline silicon with High Aggregation Degree

**Jiayang Niu [1*], Ruiling Xiao [2*], Caiwang Mao [3], Jing Xueying[1]**

[1]North China University of Science and Technology, College of Science, 063210, China;
[2]North China University of Science and Technology, College of Science, 063210, *China*;
[3]Electrical North China University of Science and Technology, Yisheng Innovation Education Base, 063210, *China*;
Corresponding author: Ruiling Xiao(niujiayang@stu.ncst.edu.cn).

*These authors contributed equally to this work and should be considered co-first authors.

**ABSTRACT** The nanoparticle size and distribution information in the SEM images of silicon crystals are generally counted by manual methods. The realization of automatic machine recognition is significant in materials science. This paper proposed a superposition partitioning image recognition method to realize automatic recognition and information statistics of silicon crystal nanoparticle SEM images. Especially for the complex and highly aggregated characteristics of silicon crystal particle size, an accurate recognition step and contour statistics method based on morphological processing are given. This method has technical reference value for the recognition of Monocrystalline silicon surface nanoparticle images under different SEM shooting conditions. Besides, it outperforms other methods in terms of recognition accuracy and algorithm efficiency.

## I. INTRODUCTION

In recent years, in the field of material science, the preparation methods of silicon crystal nanoparticles have been continuously innovated. However, so far, the identification and distribution statistics of irregular nanoparticles still need to be completed by artificial eyes, especially in the field of image identification of nanoparticles in SEM images of high-aggregation monocrystalline silicon surface coatings, there are few high-accuracy identification methods. , usually using manual annotation in professional software (such as ImageJ), but this method is very time-consuming and cumbersome, and the number of annotations is limited, the accuracy of the obtained results is lacking, and it is impossible to accurately establish the particle size distribution, quantity Effective "structure-activity relationship" between chemical properties。

Therefore, an automated technology is needed to realize automatic identification of nanomaterial structures, accurate calculation of particle area and cross-section, and quantitative statistics of their distribution. In the field of image processing, Tobia[2]proposed a histogram thresholding method based on gray similarity. This method plays a huge role in image recognition for multimodal histograms. As for the edge detection algorithm of images, as early as 2001, Lijun Dine [3] and others proposed Canny edge detection and widely used in the computer field in subsequent researches. Due to the emergence of large-scale annotated datasets such as ImageNet [7] and high-performance image processor GPUs in recent years, convolutional neural networks have great advantages in the field of image recognition [9], and are widely used in robotics, intelligent transportation, Medicine and other fields have led to the development of computer vision[4,5]. For the recognition of nanomaterials and microscopic objects by deep learning, the researchers proposed a deep learning-based deep separable convolutional U-Net network architecture [6], but due to the high cost of the current SEM image itself, and the The complexity of nanoparticle images with high aggregation properties makes the collection and construction of datasets very time-consuming and increase the workload. In addition, as the recognition objects are constantly updated and changed, the data sets required for machine learning also need to be continuously collected and reconstructed. For the recognition methods of SEM images, the existing research mainly adopts background and watershed segmentation or uses Markov random field method to segment images [8]. However, there is still a lack of research on the application of nanoparticle SEM images. The image recognition technology introduced into the microstructure analysis of nanomaterials has also developed to some extent. Lu Derong [10] comprehensively summarized the identification algorithms and macroscopic performance analysis of nanoparticles. Wang Qun [11] used histogram equalization to improve the grayscale difference between honeycombs, and adopted the widely respected OTSU method for thresholding to identify the honeycomb blocks in the image, and finally achieved the recognition by removing miscellaneous items through image morphological processing. Bremananth et al. [12] used the OTSU global threshold to qualitatively and quantitatively analyze the dispersion of TiO2 nanofillers. In this paper, a recognition process for SEM images of nanoparticles coated on Monocrystalline silicon surface is proposed by means of feature extraction, threshold processing, and image morphology processing.

### A. MOTIVATION

Since the particle identification and distribution statistics of the newly prepared nano-silicon crystal materials need to be completed by the naked eye, especially in the field of image identification of high-density nanoparticles and nanowires, there are few high-accuracy machine identification methods. This research intends to focus on this direction. In the above, using multi-threshold processing, combined mask operation technology, combined with raster scanning and linear regression, design an automatic identification system for high-density nano-images; realize automatic identification of nanomaterial structure, accurate calculation of particle area and cross-section, quantification Statistics of its distribution. It is expected to effectively realize the identification and effective segmentation of gaps between particles, provide technical support for image data for nanomaterials researchers, and study the problem of extracting microscopic data from nanometer images, study how to automatically identify nanomaterials on SEM images, and conduct research on nanomaterials. The microscopic characteristics of the image are effectively quantified, and then the relevant microscopic characteristics data are obtained.

### B. CONTRIBUTIONS

In this paper, a superimposed divide-and-conquer image recognition algorithm is proposed for the image recognition of nanoparticles in SEM images of high-aggregation monocrystalline silicon surface coatings. The main contributions of this method can be summarized as follows:

1. An identification method based on the idea of divide-and-conquer algorithm is proposed to deal with nanoparticles with high aggregation degree. The use of divide-and-conquer processing can limit the morphological processing to a single contour to be processed, effectively avoiding the problem of particle reconnection caused by morphological processing. Using the method of superimposing images, the advantages of the binarization threshold and the adaptive threshold are cleverly played.

2. The superimposed divide-and-conquer identification method in this paper has an intuitive auxiliary role in the rapid identification of the etching effect of professionals on complex SEM silicon surface images.

3. The superimposed divide-and-conquer identification method in this paper has an excellent auxiliary role in the quantitative statistics of particle size and quantity of SEM images with similar complex and high aggregation characteristics, as well as element calibration in machine learning data sets.

## II. IDENTIFICATION OBJECT AND CHARACTERISTIC ANALYSIS OF IDENTIFICATION OBJECT

### A. RECOGNITION OBJECT OF NANO PARTICLES ON MONOCRYSTALLINE SILICON COATING SURFACE

This paper will cover four kinds of SEM images of nano particles on the surface of monocrystalline silicon coating under different shooting and manufacturing conditions. Some local slices of the original images are shown in Figure 1 (a, b, c, d).

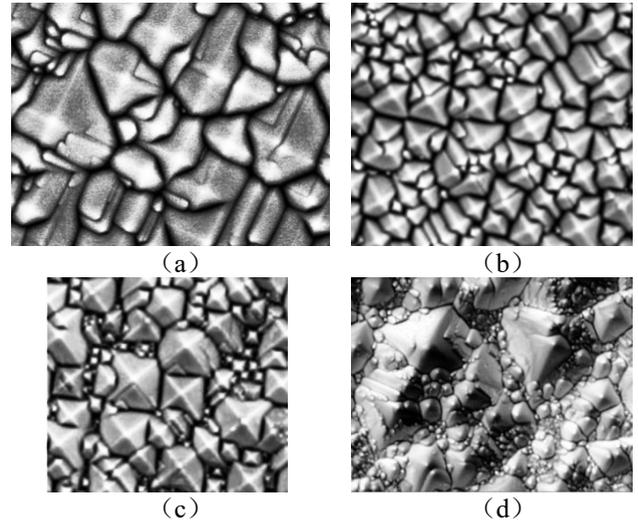

(a)　　　　　　　　(b)

(c)　　　　　　　　(d)

**FIGURE 1** SEM images of silicon crystal surface before processed

Figure 1 (a) [13] shows the SEM image of the silicon surface after corrosion in an alkaline solution at 80 C° for 25 minutes. Figure 1 (B, c) [14] shows the SEM image of the silicon surface after etching at 90 C° and 95 C° using 80 ml Di H2O, 2.9589 g KOH and 20 ml isopropanol, respectively. Figure 1 (d) [15] shows the SEM image of silicon surface etched with 15wt% koh+5wt% IPA mixed solution.

### B. FEATURE ANALYSIS OF IDENTIFIED OBJECTS:

By observing Figure 1 (a), the overall gray level of the figure is low, and the exposure of nanoparticles is relatively evenly distributed. Each nano particle can be visually distinguished through the cracks between each particle. By observing Figure. 1 (b), it can be found that the brightness of Figure. 1 (b) is higher than that of Figure. 1 (a), and the cross exposure of the center of each particle is very obvious. A single nano particle can also be well identified through the black cracks around the nano particles. Figure 1 (c) is similar to (b), but there are many irrelevant small particles around the identified object. In Figure. 1 (d), there is severe polarization exposure compared with the first three pictures, and there are many small particles around a single recognition object in the picture to form an unrelated interference item. The surrounding non particle areas also have high exposure, but the cracks between the larger nanoparticles are obvious, which can also be used as a breakthrough point for identification.

## III. ANALYSIS ON THE EFFECT OF CONVENTIONAL IDENTIFICATION METHODS AND IDENTIFICATION PROBLEMS

### A. IDENTIFY ACCORDING TO THE CHARACTERISTICS OF THE IDENTIFIED OBJECT

By directly observing Figure. 1, each nano particle can be visually distinguished by the characteristic difference between particles. After preliminary subjective analysis, such aggregated nanoparticles can be identified through the following characteristics.

For the original image, there are obvious black cracks between each particle to distinguish them. We can consider using the adaptive threshold method to enlarge these cracks so that the particles can be better identified.

From the perspective of a single nano particle, the surface of the nano particle has a higher gray level than the surrounding cracks. It can be considered to separate the particles by using equalization and threshold treatment.

From the above two basic features, according to the conventional operations of image processing [15, 16], the basic nano particle data information in the image can be obtained through "SEM image reading" - "image smoothing" - "histogram equalization" - "binary threshold processing" - "image morphological processing" - "contour detection".

### B. BASIC IDENTIFICATION STEPS
### 1) IMAGE SMOOTHING
In this paper, we use "mean filter", "Gaussian filter", "median filter" and "bilateral filter" for the four types of original images to explore the processing method for the next step recognition. Taking Figure 1(a) as an example, the filter kernel size is set to 3*3. By comparing the effects of different filtered and smoothed images, it is found that the image effects after mean filtering, Gaussian filtering and median filtering are similar. Compared with the bilateral filtering and smoothing algorithm, the noise of the image is better, and the gaps between particles are clearly distinguishable. The cracks between the nanoparticles in the image after bilateral filtering and smoothing are blurred, which is not easy to follow. In this paper, Gaussian filtering is selected for subsequent processing.

### 2) HISTOGRAM EQUALIZATION PROCESSING
Histogram equalization can effectively improve the gray level distribution of the image with too high or too low exposure, so that the image is not too bright or too dark. In order to meet the requirements for threshold selection in subsequent experiments, it is necessary to perform histogram equalization processing on the image. Figure. 2 is a partial slice of Figure. 1 after histogram equalization.

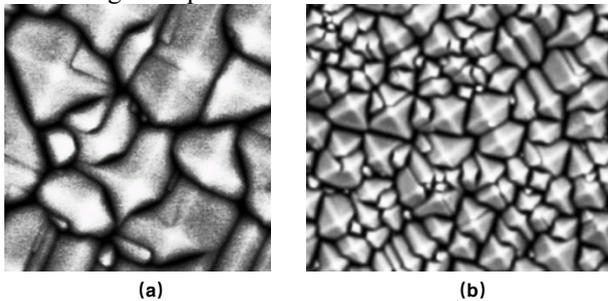

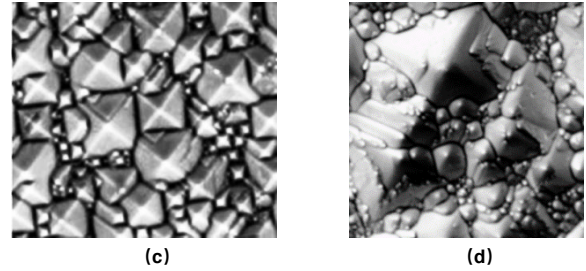

FIGURE 2 Partial sections after histogram equalization

By comparing the four types of images with different properties in Figure 2, the single recognition object that was dark before in Figure 2(a) is obviously brightened, and the surrounding cracks are darker. In Figure. 2(b, c), the central cross bright area with high exposure itself becomes brighter, and the dark area outside the single cross area of the recognition object becomes relatively darker. Figure 2(d) The exposure is brighter on one side and darker on the other side.

### 3) IMAGE THRESHOLDING
According to the properties of dense nanoparticles, it is relatively easy to segment them through the black gaps between them. This paper uses binarization threshold processing, and its threshold setting should be selected at a gray level between black and white. By observing Figure 2, according to the gray level distribution of various images, the gray value of the boundary between the dark area and the bright area is selected as the threshold for threshold processing. The partial slice of the image after binarization thresholding is shown in Figure 3. (Thresh: a = 100, b = 80, c = 60, d = 50 ).

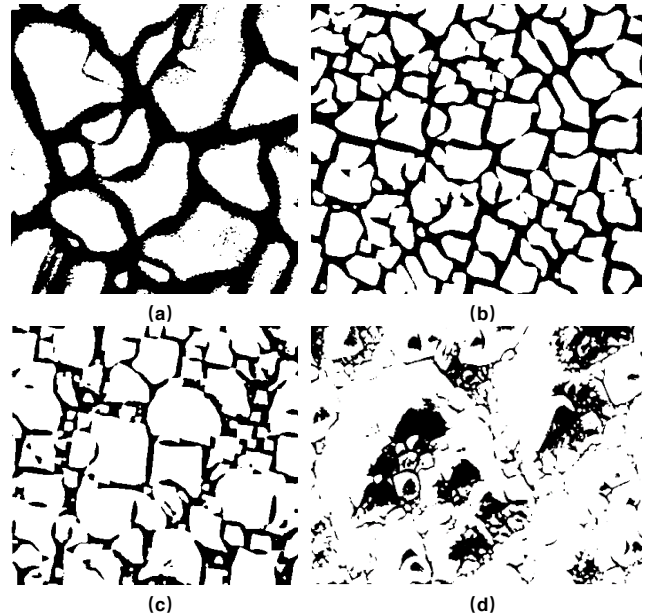

FIGURE 3 Partial sections after threshold processing

By comparing the images of these four different types of recognition objects after binarization threshold processing, it is found that the binarization threshold processing effect in Figure 3(a) is better, and the recognition objects are all perfectly processed by thresholding. Segmentation has been performed, however, observe Figure 3(b,c), the recognition object monomers in the picture are not perfectly separated by binarization threshold processing, there are still small connecting lines between each nanoparticle, and the original image Mid-to-dark nanoparticles are filtered out by the threshold. In addition, there are many small white spots at the edges of the particles, and many small black spots in the center of the particles. However, looking at Figure 3(d), it is found that the image has serious recognition problems. Due to its own uneven exposure, thresholding can only identify half of the large nanoparticles. Therefore, morphological processing of the pictures is needed to help further identification.

4) IMAGE MORPHOLOGICAL PROCESSING

Morphological processing includes image erosion, image expansion, opening operation and closing operation. Through image morphological operations, the redundant or vacant parts in the image can usually be processed, the irrelevant pixels in the image can be removed, and the vacancies in the image can be filled, thereby helping image recognition [17]. The principle is to scan from the upper right corner element of the image to the lower left corner element of the image through a structure element (kernel) of a specified size, from left to right, and from top to bottom. When the elements of the struct overlap, perform a specified morphological transformation on the elements in the image covered by all elements in the structuring element. The following takes Figure 3(a) as an example to perform morphological processing.

*Open operation processing:*

By performing the opening operation on the image (corrosion first and then expansion), the 3*3 structural elements are used to eliminate the small white spots outside the nanoparticles, and at the same time, two particles with only very small line segments in the middle are connected. Particulate matter is divided into two particles. The generated image is shown in Figure 4(a). It can be seen that the small white spots outside the nanoparticles are basically eliminated, and the edges of the particles are relatively smoothed.

*Close operation processing:*

Set the morphological processing method of the closing operation, construct a 3*3 structure element, and iterate a closing operation on the image. Through the closing operation of the image (expansion first and then erosion), the small points inside the particles are eliminated. reduce. After this step, the segmentation between nanoparticles by conventional means has been basically completed.

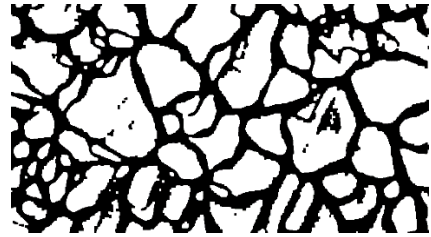

FIGURE 4 (a) Image after opening operation

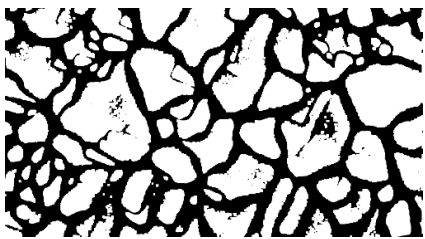

FIGURE 5 (a) Image after closed operation

In order to use the morphological processing method, the threshold size of the threshold value processing in Figure 3(d) is increased here, and the local slices of the b, c, d images after the morphological processing after increasing the threshold value are shown in Figure 6(b, c, d) .

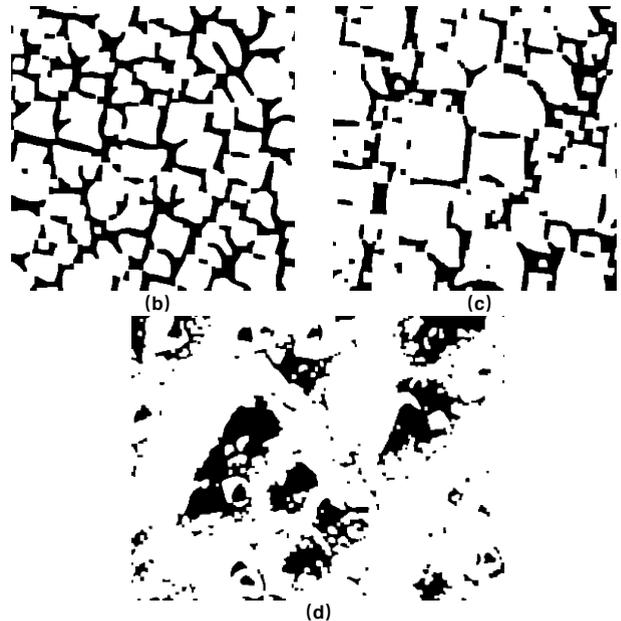

FIGURE 6 Images after morphological processing

Through the comparison, it is found that there are still some connections between the nanoparticles in Figure 6(b) that are difficult to remove, and even the particles that were not originally connected are reconnected. While the original multiple nanoparticles in Figure. 6(c) are merged into one large nanoparticle, Figure. 6(d) cannot even identify a single nanoparticle.

5) RECOGNITION OBJECT CONTOUR PROCESSING

In order to extract all the particles in the image in sequence, this paper uses the contour recognition algorithm provided by the OpenCv library to identify and count the contour information array of each nanoparticle (white block) with a single-layer outer contour recognition pattern. Through the contour drawing function, the contour of the image is drawn in the original image according to each boundary point of the identified contour information array. Since the outer contour is the recognition object in this contour recognition mode, the inner contour is automatically ignored, so the influence of black spots inside the nanoparticles is not considered. In this way, contour processing is performed on the four types of morphologically processed images, and the recognition results obtained are shown in Figure 7. In this paper, the detected particles on the surface of nanomaterial Monocrystalline silicon are marked with red outlines.

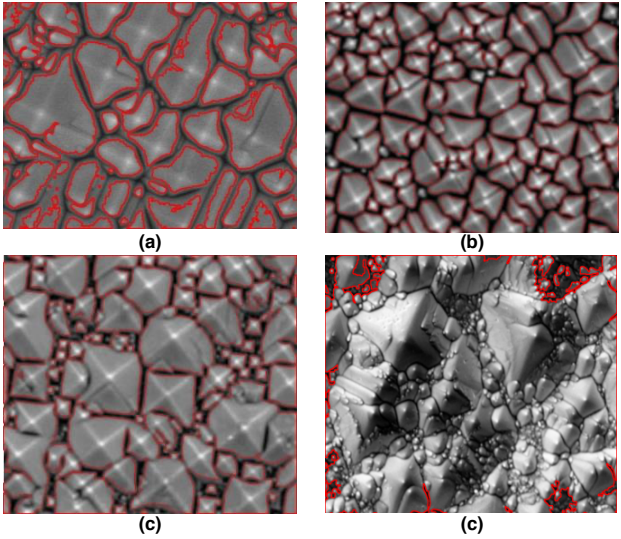

**FIGURE 7** Images after contour processing

Through comparison, picture a is better for identification by conventional means, while picture b and c are still connected to each other between nanoparticles, and picture d nanoparticles are directly identified as one piece, and there is serious polarization identification sex.

## IV. IDENTIFY KEY ISSUES AND ANALYZE

The following difficulties are summarized in the identification of four types of SEM images of Monocrystalline silicon surface coatings with different properties by conventional means.

Difficulties of high local exposure of particles: By observing Figures b, c, and d, it can be found that these three types of images have local high exposure, and errors are prone to occur in various threshold processing.

### 1) DIFFICULTIES IN THE SELECTION OF THE BINARIZATION THRESHOLD

Observe Figure 2, and we can find that all the pictures after histogram equalization do not have obvious two extreme gray-level clusters. If the threshold is chosen too small, the gaps between the individual nanoparticles will not be fully utilized to segment each other, while if the threshold is chosen too large, some underexposed nanoparticles or nanoparticles will be partially removed, resulting in Recognition integrity declines.

### 2) DIFFICULTIES DUE TO TOO DENSE PARTICLES

From Figure d, it can be found that some nanoparticles are too dense, and the grayscale of the boundary cracks is not much different from itself, so it is difficult to separate them through a general threshold.

### 3) DIFFICULTIES IN THE INTERFERENCE OF IRRELEVANT TINY PARTICLES

Observe all the recognition results, and contour recognition also circles many irrelevant tiny particles

### 4) THE TRADE-OFF PROBLEM OF GAP FILLING CAUSED BY GAP RECONNECTION

In actual recognition, there are often gaps in the recognized object. If you want to use the closing operation (expansion and then corrosion) to fill the gap, the originally disconnected gaps are connected. It is then difficult to choose whether or not to fill the void at the risk of reconnecting an otherwise disconnected gap.

## V. ADAPTIVE THRESHOLD PROCESSING METHOD AND IDENTIFICATION DIFFICULTY ANALYSIS

### A. ADAPTIVE THRESHOLDING TO IDENTIFY INTERPARTICLE GAPS

Adaptive thresholding can perform better thresholding on images with uneven exposure distribution. It obtains a threshold by calculating the weighted average of the adjacent areas around each pixel, and subtracts a certain difference from the threshold. Threshold to process the current pixel. Its scanning convolution kernel has "average convolution kernel" and "Gaussian convolution kernel". Assuming that the pixel value of the pixel around the center point is $P(x, y)$, the weight ratio of this point relative to the entire convolution kernel is $W(x, y)$, and the difference between the two is is $D$, the adaptive threshold formula is:

$$F_T(x, y) = -D + \sum P(x_i, y_i) W(x_i, y_i)$$

Whether the final center point is 255 (white) or 0 (black) depends on its size comparison with the calculated threshold:

$$T = \begin{cases} 255 \ if \ \ P(x_0, y_0) > P_T(x_0, y_0) \\ 0 \ \ \ if \ \ P(x_0, y_0) \leq P_T(x_0, y_0) \end{cases}$$

Under the influence of noise, generally speaking, the larger the $D$ value, the less affected by the noise, but at the same time the recognition effect will also be reduced; on the contrary, the smaller the $D$ value, the greater the effect of the noise, and the recognition effect will be relatively good. promote.

In order to solve various problems in routine operation, according to the identification characteristics of adaptive threshold, after image preprocessing, adaptive threshold processing should be used instead of binarization threshold processing, and by identifying the cracks between nanoparticles, each nanoparticle can be segmented. particle effect.

## B. ANALYSIS OF THE EFFECT OF SELF-ADAPTIVE THRESHOLD PROCESSING TO IDENTIFY PARTICLE GAPS

Figure 8 shows the local slices of the renderings obtained by using adaptive thresholds for image processing of Figure 3 (a, b, c, d) that have been preprocessed.

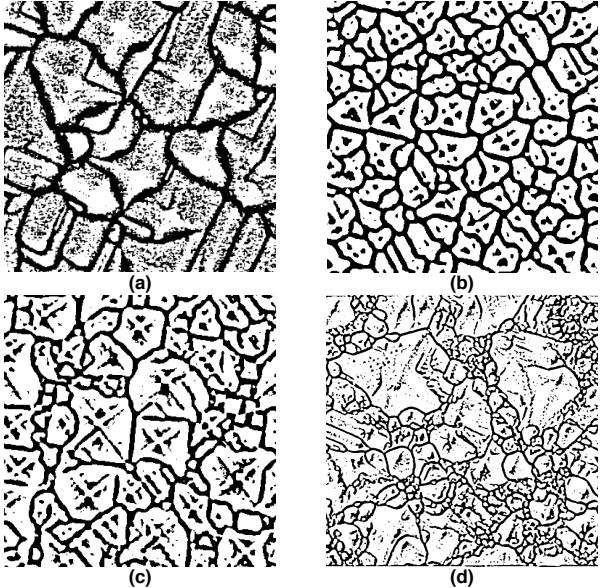

**FIGURE 8** Partial sections after adaptive threshold processing

By comparing the recognition effects of each image, it is not difficult to find that the recognized images do not achieve the expected effect. Although the cracks between the individual nanoparticles were well identified, some tiny cracks and uneven exposure of the individual nanoparticles themselves led to misidentification by adaptive thresholding. In Figure. 9(a), the surface of the identified nanoparticle is covered with black spots due to the noise effect. In Figure. 9(b, c, d), the adaptive threshold processing is wrongly determined to be a crack due to the cross-shaped exposure and side polarization of the surface.

## C. DIFFICULTIES IN USING ADAPTIVE THRESHOLD PROCESSING AND SUBSEQUENT OPERATIONS

### 1) CROSS-TYPE EXPOSURE AND SURFACE CRACKS
When using adaptive processing, cross-type exposure and surface cracks are easily misidentified as cracks between particles by the computer, resulting in some nanoparticles being cut into multiple small particles, extremely greatly affect the recognition effect.

### 2) CARRIER CRACKS
The use of adaptive processing is that the cracks in the carrier will also be misjudged as the cracks between the particles, so that the areas that are not originally regarded as nanoparticles are segmented and misjudged as nanoparticles.

### 3) THE TRADE-OFF PROBLEM OF PARTICLE SPLITTING CAUSED BY THE EXPANSION OF CRACKS

The adaptive threshold processing cannot perfectly achieve the effect of dividing all nanoparticles. In order to expand the cracks between the particles, the morphological opening operation should be used for processing. However, too small structural elements have poor effect of expanding cracks, and too large structural elements can easily lead to the splitting of complete particles.

## VI. SUPERPOSITION DIVIDE AND CONQUER IDENTIFICATION METHOD

### A. ALGORITHM DESIGN BACKGROUND
In view of the above problems and difficulties, the algorithm needs to solve the problem of identifying nanoparticles in SEM images of high-density Monocrystalline silicon surface coatings. It is necessary to meet the requirements as much as possible that even in the case of uneven exposure of nanoparticles, threshold processing can be used to segment each closely connected nanoparticle, and at the same time, it is necessary to avoid the irrelevant interference generated in the identification of circles when using contour processing. Support material for small particles and nanoparticles. In order to meet the needs of researchers to better study the macroscopic physical properties of the material through the statistical data of the microscopic properties of the Monocrystalline silicon surface obtained by image recognition

### B. ALGORITHM PRINCIPLE
Through the identification results obtained by the conventional identification algorithm and adaptive threshold processing, it is found that the conventional binarization threshold processing can well avoid the cracks in the center of the particles, but it is difficult to separate the particles through the gaps between the particles. Threshold processing can well segment each particle through the gap between particles, but it is difficult to avoid false segmentation caused by cross-shaped or half-edge exposure in the center of the particle.

The final recognition result also often has a large number of incomplete segmentation of particles or excessive segmentation of particles. Excessive use of morphological processing will make the processing results worse and worse. If each preliminary segmented block can be extracted separately for processing. Processing, without affecting other particles with better segmentation effect, can further improve the recognition effect.

Based on the phenomena found in the above recognition experiments, this paper proposes a "superposition divide and conquer recognition algorithm". Each recognition block is extracted separately for further processing, and the number of iterations is increased according to the recognition effect. The specific algorithm steps are as follows:

| Algorithm: Superposition divide and conquer algorithm |
| --- |
| **Input** : original graph data matrix $P-$ |
| **Output** : silhouette collection $Sc+$ |
| **Step1:** After preprocessing $P-$, the graph data matrix P is |

obtained.

**Step2:** The graph data matrices $P_{a-}$ and $P_{b-}$ are obtained by using adaptive threshold operation and binarization threshold operation respectively for $P$.

**Step3:** Perform the opening operation on $p_{a-}$ and $p_{b-}$ respectively to expand the crack effect to obtain $P_a$ and $P_b$.

**Step4:** Bitwise AND the matrices $P_a$ and $P_b$, $P_m = P_a \& P_b$;

**Step5:** Perform the first round of contour recognition to get the contour set, $S_{c-} = Contours(P_m)$;

**Step6:** Get a new contour set by removing the minimal area contour elements $S_c$。

**Step7:** Extract each element $c_i$ in the new contour set separately, and press the coordinate information set in $c_i$, in an all-zeros matrix of the same size as matrix $P$, Change the 0 of the corresponding coordinate position to 255, and then write all the closed loop areas composed of 255 in the matrix as 255, and get the matrix $P_{ci}$ composed of a single identification block.

**Step8:** Perform a closing operation on the matrix data alone to get $P_{ci}+$, Again preset the same size all 0 matrix $P+$, Iteratively bit-OR's all individual matrix data with $P+$, $P+ = P | P_{ci}+$;

**Step9:** Finally, perform contour recognition $Sc+ = Contours(P+)$, The final recognition result contour set $Sc+$ can be obtained。

## VII. APPLICATION OF SUPERPOSITION DIVIDE AND CONQUER RECOGNITION ALGORITHM

### A. ENLARGEMENT OF PARTICLE GAP EFFECT

The smoothing processing and histogram equalization required in the early preprocessing are consistent with the conventional processing, and the threshold processing is unchanged. Here, the conventional means are directly used to obtain the binarized threshold processing results in Figure 3 and the adaptive threshold processing results in Figure 8. For the two sets of result graphs obtained, the particle gap effect is enlarged by using the open operation, so that the nanoparticles that are not completely separated can be further separated. Although the cracks inside the particles will also expand at this time, it has little effect on subsequent operations. Part of the effect diagram after expanding the gap is shown in Figure 9.

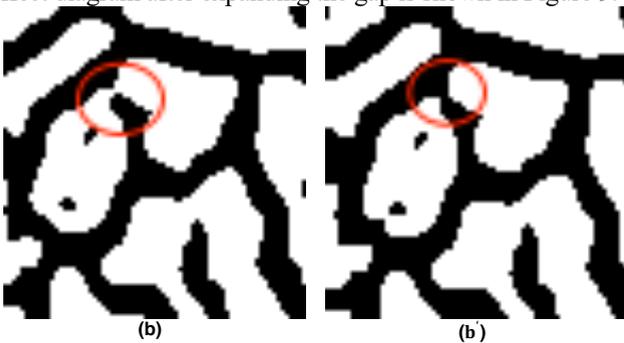

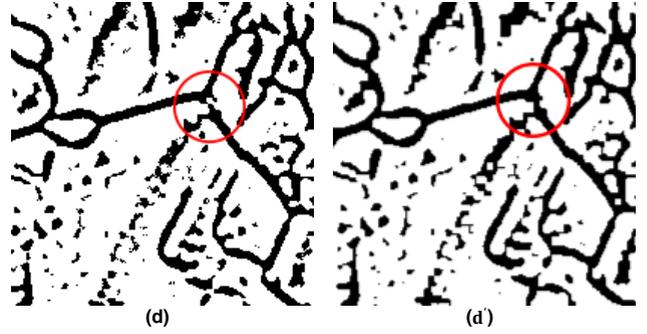

**FIGURE 9** Comparison of before and after gap widening

Observing Figure. 9(b,b') and Figure. 9(d,d'), it can be found that the nanoparticles that still have tiny connections are now completely separated. The whole image is more fragmented, which is of great significance for later screening of too dense irrelevant tiny particles.

### B. IMAGE OVERLAY

By superimposing the image obtained by adaptive thresholding with the gap effect expansion processing and the image obtained by binarization thresholding, it is very easy to face the large unrelated tiny particle interference area or non-grain dark area in the original image. It has a strong filtering effect [18]. After stacking, the large-particle gap image blocks obtained by adaptive thresholding can be retained at the same time, and large irrelevant small-particle blocks can be directly removed. By superimposing Figure. $10(c'')$ and $10(c')$, the obtained effect is shown in Figure. $10(c)$.

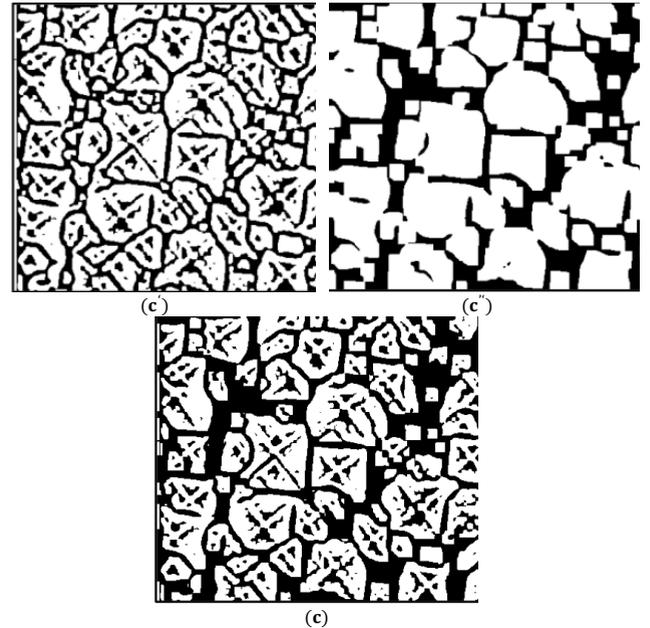

**FIGURE 10** Superposed by AND operation

By comparing the three pictures in Figure 10, it can be clearly found that Figure 10(c) after superimposed with the operation not only retains the better particle gap segmentation effect brought by the adaptive threshold processing, but also makes a good use of the two The thresholded image filters out large patches of irrelevant small particles.

### C. DE-MINIMAL AREA CONTOUR PROCESSING
The extremely small area contour here is relative to the large particles in the original image. In order to further filter out irrelevant small particles, it is necessary to calculate the contour of each element $c_{i-}$ in the contour set $S_{c-}$ by computer. By setting an area screening threshold relative to large particles, the computer automatically filters out the contour data of particles with extremely small areas according to the threshold. Since the contour data information of each particle wraps the outermost contour of each particle and does not include the inner contour, the internal voids can be ignored directly. Taking the d image as an example, the image processed by removing the minimal area contour is shown in Figure 11.

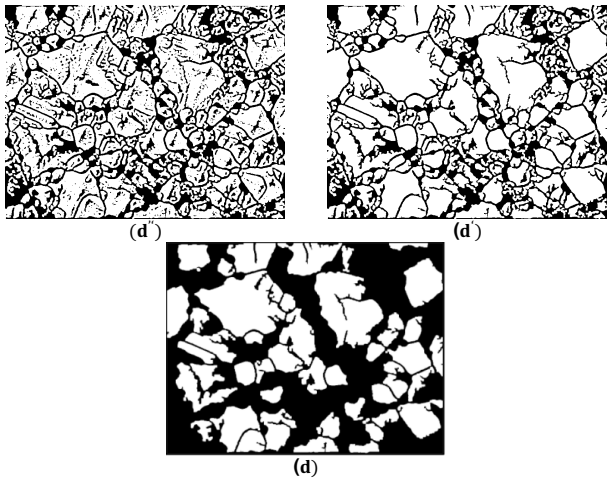

**FIGURE 11** Filter out the contour which acreage is small

Observing Figure. 11, it can be found that a lot of irrelevant small particles are filtered out, and a set S_c that only leaves the main identification particle profile information is obtained. The filtering of very small area contours greatly enhances the correctness of particle data statistics in the later stage.

### D. DIVIDE AND CONQUER
Divide-and-conquer processing is mainly used to solve the situation where most of the internal cracks of large particles lead to complete cracking or partial cracking of the particles themselves in the previous series of processing. Normally, although the cracks can be repaired by using the closed operation process on the whole picture, due to each The individual particles are very close to each other. Using closed arithmetic morphological processing on the whole image will cause the separated nanoparticles to re-aggregate. Therefore, the whole image needs to be divided and conquered, and each particle needs to be divided and conquered. Extracted separately for crack repair work. The contour information elements in the contour set Sc are sequentially extracted by the computer, and individually drawn in an all-zero matrix of the same size as the original image data matrix, and the interior of the contour closed loop is filled. To achieve the purpose of repairing cracks and gaps. Taking images c and d as an example, the images after dividing and conquering are shown in Figure 12 and Figure 13.

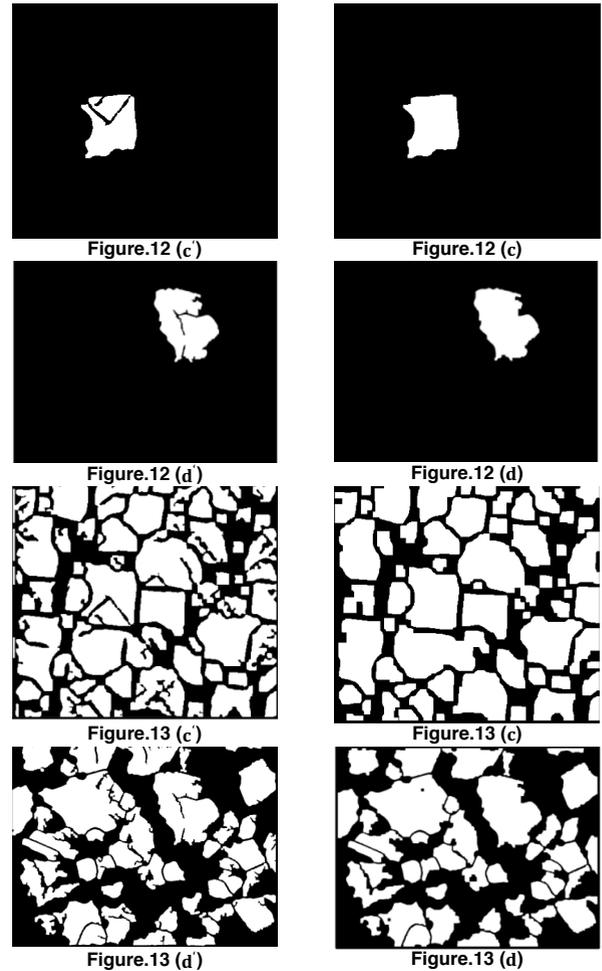

**FIGURE 12 13** The result and procession from divide-and-conquer processing

Observing Figures 12 and 13, it can be found that the particles that caused partial or overall cracking in the previous operation have completely repaired their own cracks by dividing and conquering, while the original adjacent particles that are very similar and have no contour interconnection are in the dividing and conquering operation. The problem of reconnection caused by the overall closing operation is completely avoided.

### E. ANALYSIS OF IDENTIFICATION RESULTS
By using the superposition divide-and-conquer algorithm to identify the SEM images of the four types of Monocrystalline silicon surface coating nanoparticles, the identified particle outlines can be marked to obtain a set of images as shown in

Figure 14. The SEM-like images obtained in Figure 7 are compared.

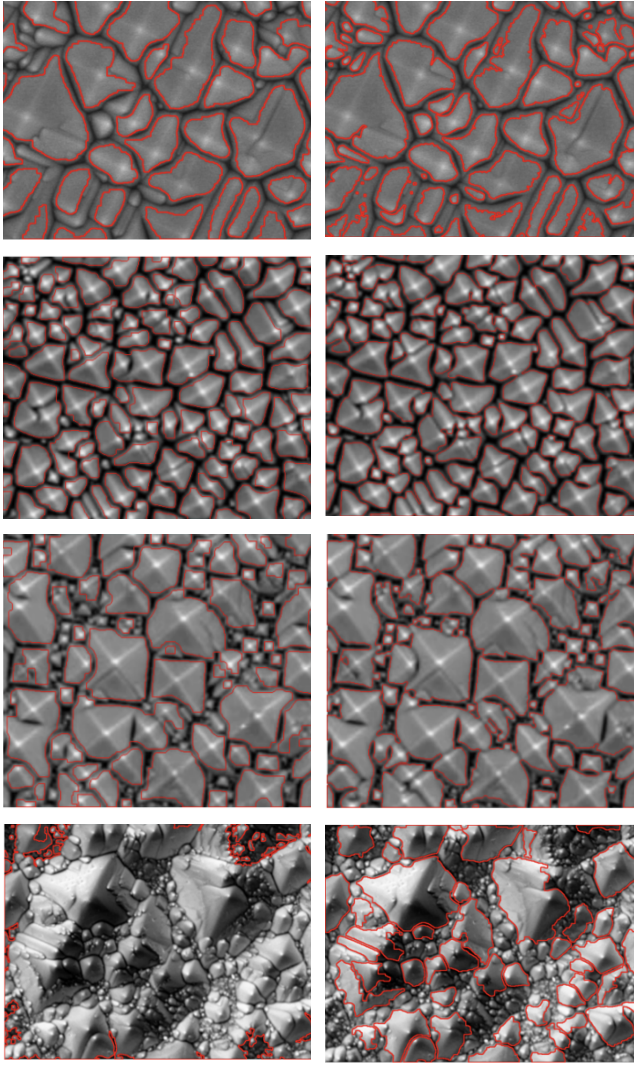

**FIGURE 7 (a,b,c,d)**
Conventional method

**FIGURE14(a,b,c,d)**
Superposition and divide-and-conquer method

By comparing Figure. 7(a) and Figure. 14(a), we can find that the segmentation and recognition effect of high-aggregation nanoparticles in Figure. a is basically the same using the conventional method and the superposition divide-and-conquer algorithm. Figure 7(b, c) and Figure 14(b, c), the superposition divide-and-conquer method is more advantageous than the conventional method in dealing with tighter nanoparticle segmentation. Finally, observe and compare Figure. 7(d) and Figure. 14(d), in the face of such a large number of irregular particles and excessively polarized SEM images, the superimposed divide-and-conquer algorithm obtained an excellent recognition effect compared with the conventional method. Nanoparticles, which had been completely recognisable in confusion, were segmented and labelled perfectly.

## VIII. CONCLUSION

Through comparative experiments, this paper discusses how to better and effectively identify the SEM images of nanoparticles coated on the surface of Monocrystalline silicon with high aggregation degree when the particles are arranged more densely and the surface is rougher. This method abandons the one-line identification of the traditional method, and solves the problem of difficult identification of nanoparticles with high aggregation degree by combining the method of image superposition and particle divide and conquer processing. In addition, the processing results of the high-aggregation-degree Monocrystalline silicon SEM images by the method in this paper can more intuitively reflect the etching effect of the silicon surface during the experiment, and also have advantages in image recognition and quantitative statistics for other high-aggregation-degree nanoparticles. Reference meaning. However, the identification method in this paper mainly relies on the grayscale change of the image to identify the particles. When facing the particle gap with almost no change in grayscale, the identification method is not effective for the identification of single particles. In the future, the existing research results will be further combined with the image recognition model in machine learning, and the model matching method will be used to enhance the identification and segmentation of particles under the preprocessing of superimposed divide and conquer.